\documentclass[acmsmall]{acmart}
\AtBeginDocument{%
  }

\setcopyright{acmlicensed}
\copyrightyear{2018}
\acmYear{2018}
\acmDOI{XXXXXXX.XXXXXXX}
\acmConference[SIGSPATIAL 2025]{November 03--06,
  2025}{Minneapolis, MN, USA}

\acmISBN{978-1-4503-XXXX-X/2018/06}

\usepackage[utf8]{inputenc}
\usepackage{newunicodechar}
\newunicodechar{₁}{$_1$}
\usepackage{graphicx}   
\usepackage{caption}    
\usepackage{subcaption} 

\begin{document}

\title{ST-GraphNet: A Spatio-Temporal Graph Neural Network for Understanding and Predicting Automated Vehicle Crash Severity}

\author{Mahmuda Sultana Mimi}
\orcid{0009-0007-8534-3633}
\affiliation{%
  \institution{Texas State University}
  \streetaddress{601 University Drive}
  \city{San Marcos}
  \state{Texas}
  \country{USA}
  \postcode{78666}
}
\email{qnb9@txstate.edu}
 
\author{Md Monzurul Islam}
\orcid{0009-0007-3670-6100}
\affiliation{%
  \institution{Texas State University}
  \streetaddress{601 University Drive}
  \city{San Marcos}
  \state{Texas}
  \country{USA}
  \postcode{78666}
}
\email{monzurul@txstate.edu}
 
\author{Anannya Ghosh Tusti}
\orcid{0009-0001-5102-5182}
\affiliation{%
  \institution{Texas State University}
  \streetaddress{601 University Drive}
  \city{San Marcos}
  \state{Texas}
  \country{USA}
  \postcode{78666}
}
\email{gpk30@txstate.edu}
 
\author{Shriyank Somvanshi}
\orcid{0009-0008-3723-0607}
\affiliation{%
  \institution{Texas State University}
  \streetaddress{601 University Drive}
  \city{San Marcos}
  \state{Texas}
  \country{USA}
  \postcode{78666}
}
\email{shriyank@txstate.edu}
 
\author{Subasish Das}
\orcid{0000-0002-1671-2753}
\affiliation{%
  \institution{Texas State University}
  \streetaddress{601 University Drive}
  \city{San Marcos}
  \state{Texas}
  \country{USA}
  \postcode{78666}
}
\email{subasish@txstate.edu}


\begin{abstract}
Understanding the spatial and temporal dynamics of automated vehicle (AV) crash severity is critical for advancing urban mobility safety and infrastructure planning. In this work, we introduce \textbf{ST-GraphNet}, a spatio-temporal graph neural network framework designed to model and predict AV crash severity by using both fine-grained and region-aggregated spatial graphs. Using a balanced dataset of 2,352 real-world AV-related crash reports from Texas (2024), including geospatial coordinates, crash timestamps, SAE automation levels, and narrative descriptions, we construct two complementary graph representations: (1) a fine-grained graph with individual crash events as nodes, where edges are defined via spatio-temporal proximity; and (2) a coarse-grained graph where crashes are aggregated into Hexagonal Hierarchical Spatial Indexing (H3)-based spatial cells, connected through hexagonal adjacency. Each node in the graph is enriched with multimodal data, including semantic, spatial, and temporal attributes, including textual embeddings from crash narratives using a pretrained Sentence-BERT model. We evaluate various graph neural network (GNN) architectures, such as Graph Convolutional Networks (GCN), Graph Attention Networks (GAT), and Dynamic Spatio-Temporal GCN (DSTGCN), to classify crash severity and predict high-risk regions. Our proposed ST-GraphNet, which utilizes a DSTGCN backbone on the coarse-grained H3 graph, achieves a test accuracy of 97.74\%, substantially outperforming the best fine-grained model (64.7\% test accuracy). These findings highlight the effectiveness of spatial aggregation, dynamic message passing, and multi-modal feature integration in capturing the complex spatio-temporal patterns underlying AV crash severity.
\end{abstract}



\received{5 June 2025}
\received[revised]{XX}
\received[accepted]{XX}

\maketitle

\section{Introduction}
Crashes involving automated vehicles (AVs) have become increasingly visible as these systems are tested and deployed across public roadways. In California alone, as of May 2025, over 821 AV-related crashes were reported, including more than 200 at intersections \cite{californiadmv2023}. Among these, rear-end collisions where AVs were struck by conventional vehicles accounted for over half of all intersection crashes. Despite being engineered to eliminate human error, which is implicated in over 90 percent of crashes in the United States \cite{singh2015critical}, AVs have not yet demonstrated unequivocal safety benefits across all operational conditions. These early crash patterns suggest that additional insight is needed to understand when and why AVs are vulnerable, particularly in mixed traffic environments and complex road geometries \cite{song2022intersection, arvin2020safety, mahdinia2021integration, hossain2024potential, he2021introducing}.

Traditional modeling approaches, such as statistical regression or conventional machine learning, have been applied to AV crash data to estimate injury severity or identify contributing factors. Although informative, these methods often assume linearity or independence among input features and do not fully capture the sequential and spatial complexity surrounding AV operations \cite{boggs2020exploratory,chen2021deep,chen2021novel}. Moreover, these models typically fail to represent the unique operational logic of AVs, such as their reliance on environmental sensing, pre-programmed responses, and constrained operational design domains. More recent studies have adopted a scenario-based approach, analyzing vehicle movements and behaviors immediately before a crash. For example, failure to adequately decelerate or misinterpreting traffic signals has been observed in a significant number of intersection-related AV crashes, indicating a need for behavior-aware prediction systems \cite{liu2024social,hong2024knowledge}.

Graph neural networks (GNNs) have emerged as a promising framework for modeling the dynamic and relational nature of traffic systems. These models represent entities such as vehicles or road segments as nodes within a graph, enabling the learning of complex interactions across space and time. A notable contribution in this domain is the Dynamix Spatio-Temporal Graph Convolutional Network (DSTGCN), which successfully integrated spatial and temporal dependencies to predict crash occurrences at the road-segment level with greater precision than traditional models \cite{yu2021deep}. Building on this, the Sequential Spatio-Temporal Graph Convolutional Network (SST-GCN) combined temporal memory units with graph-based spatial learning to capture accident patterns at a minute-level resolution \cite{kim2024sst}.

Additional advancements have utilized real-time traffic data collected by unmanned aerial vehicles to support spatio-temporal modeling. For instance, the integration of adaptive adjacency matrices and dynamic graph structures in recent models has enabled robust prediction of traffic flow under non-stationary urban conditions \cite{ma2024spatio}. These graph-based architectures have significantly improved the ability to model complex roadway systems. However, their application to AV-specific crash data remains limited. Existing models often generalize across vehicle types without accounting for AV sensor behavior, fallback mechanisms, or human-machine interaction in partially automated systems \cite{xu2020automation, ren2023vehicle, liu2024social, yang2021secure, van2019real, martin2019identification}.

The present study aims to address these limitations by developing a spatio-temporal graph neural network framework, referred to as ST-GraphNet, customized for the characteristics of AV-involved crashes. The proposed model embeds crash events within a dynamic graph structure informed by temporal sequences, roadway topology, and vehicle-environment interactions. The objectives are threefold. First, to analyze real-world AV crash data and extract temporal and spatial risk patterns. Second, to evaluate the performance of spatio-temporal GNNs against baseline models in severity prediction tasks. Third, to identify influential features associated with elevated crash severity in automated driving systems. The following research questions guide this study: To what extent do spatio-temporal dependencies improve the prediction of AV crash severity? Which behavioral, environmental, and vehicular features most influence injury outcomes? How can graph neural networks be adapted to represent operational dynamics unique to AV systems?

\section{Related works}

\subsection{Crash Severity Modeling in Transportation }
Traditional crash severity and frequency models have long relied on statistical count and regression techniques, primarily Poisson and Negative Binomial (NB) formulations to relate crash outcomes to roadway and traffic factors. Poisson models could estimate crash frequencies as a function of traffic volume, geometric attributes, and control measures:
\begin{equation}
    \lambda_{i}=\exp \left(\beta_{0}+\beta_{1} x_{i 1}+\beta_{2} x_{i 2}+\cdots+\beta_{p} x_{i p}\right),
\end{equation}
where each $x_{ij}$ is a roadway or traffic-related covariate (e.g., average annual daily traffic, speed limit, lane width) \cite{wang2017multivariate}. Because observed crash counts often exhibit overdispersion, Negative Binomial approaches introduced a dispersion parameter $\alpha$, modifying the variance structure to $\mathrm{Var}(Y_i) = \lambda_i + \alpha\,\lambda_i^2$ to better fit real-world crash data \cite{lord2010statistical}. Although effective for many applications, Poisson and NB models depend on carefully hand-crafted covariates and assume a log-linear relationship between predictors and crash frequency. In practice, roadway risk factors often interact nonlinearly, especially under varying environmental and traffic conditions \cite{carrodano2024data}. Thus, linear link functions can misrepresent true relationships.

Spatially explicit regression approaches—such as Conditional Autoregressive (CAR) or Simultaneous Autoregressive (SAR) models explicitly model spatial autocorrelation by including a neighborhood matrix $W$, where $w_{ij} = 1$ if segments $i$ and $j$ are adjacent and 0 otherwise. A spatial lag model for crash counts may be written:
\begin{equation}
    Y_{i}=\rho \sum_{j} w_{i j} Y_{j}+X_{i} \beta+\epsilon_{i}, \quad \epsilon_{i} \sim N\left(0, \sigma^{2}\right),
\end{equation}
where $\rho$ captures the strength of spatial dependence. Similarly, a spatial lag NB model extends the NB framework with a spatial lag term. Although these methods improve upon purely aspatial GLMs by accounting for spatial clustering, they require a priori definition of the spatial weights matrix 
$W$. They also impose linear relationships between covariates and the logarithm of crash rates, and they often ignore temporal dynamics or treat time via separate fixed-effects dummy variables \cite{anselin2009spatial}. A geographically weighted Poisson regression (GWPR) applied to county-level crash data found that modeling spatially varying coefficients reduced mean absolute deviation (MAD) by 23.42 \% and mean squared prediction error (MSPE) by 66.11 \% compared to a GPM, capturing nonstationary relationships between built environment and crash counts \cite{li2013using}.


Traditional statistical methods for crash modeling face several key limitations. They rely heavily on handcrafted features and assume linear relationships and specific error distributions, which can miss complex, nonlinear patterns in crash‐severity mechanisms. Spatial and temporal dependencies are often modeled separately, requiring complex hierarchical or Bayesian frameworks to integrate both, which increases the computational burden. These approaches also struggle to incorporate high‐dimensional or unstructured data (such as crash narratives or irregular road‐network topologies) without extensive preprocessing. As the number of segments, time points, and covariates grows, fitting models like GLMMs or Bayesian spatial regressions becomes increasingly intractable, limiting their real‐time applicability.

\subsection{Deep Learning Approaches for Spatio-Temporal Data}
Deep learning has emerged as a promising alternative for modeling complex spatio‐temporal phenomena, leveraging hierarchical feature extraction to automatically learn from raw inputs. In transportation safety and traffic forecasting, Convolutional Neural Networks (CNNs) and Recurrent Neural Networks (RNNs), particularly Long Short‐Term Memory (LSTM) and Gated Recurrent Unit (GRU) variants, have been widely adopted. For instance, Lv et al. \cite{lv2014traffic} proposed a deep LSTM (Stack Autoencoder Model) to predict traffic flow on highway segments using time, series traffic flow data, achieving substantial improvements over linear regression (BP NN, RW and SVM). In another study, T-LSTM achieved an RMSE of 50.54 (MAPE 6.09 \%), the lowest among all tested models (T-GRU, SAE, DBN, FFNN, SVM, KNN, and ARIMA), demonstrating superior short-term traffic flow prediction accuracy and providing improved temporal inputs that can enhance crash severity prediction \cite{mou2019t}.The RNN‐based models can learn temporal dependencies (e.g., rush‐hour surges), but they often treat spatial context implicitly (by concatenating neighboring features) rather than modeling it explicitly.

To better capture spatio‐temporal correlations, researchers have combined convolutional layers with recurrent architectures. ConvLSTM extends standard LSTM by replacing matrix multiplications in the input‐to‐state and state‐to‐state transitions with convolutional operations \cite{xingjian2015convolutional}. This structure preserves spatial locality and captures short‐term temporal evolution in gridded data (e.g., traffic speed maps). In one study, Hetero-ConvLSTM model is proposed which showed the lowest RMSE score (0.021) compared to standard ConvLSTM (RMSE: 0.117) and lifted prediction accuracy to 87.5 \%, significantly outperforming all baselines \cite{yuan2018hetero}. Likewise, 3D CNNs have been used to simultaneously model spatial neighborhoods and temporal “slices” \cite{tran2015learning}. 


\subsection{Graph Neural Networks in Transportation Safety}
Graph Neural Networks (GNNs) have gained significant attention in intelligent transportation systems (ITS) due to their ability to model non-Euclidean spatial relationships and capture dynamic temporal dependencies in road networks. One of the foundational GNN models, the Graph Convolutional Network (GCN) proposed by Kipf and Welling \cite{kipf2016semi}, introduced node-level message passing for classification tasks and laid the groundwork for subsequent spatial graph modeling. Extensions such as the Graph Attention Network (GAT) improved on this by learning weighted edge importance through attention mechanisms, enhancing interpretability and adaptability in dynamic traffic conditions \cite{gao2024uncertainty}. Similarly, GraphSAGE has proven effective in sparse transportation networks by enabling inductive learning across unseen nodes through neighborhood sampling and aggregation, particularly for traffic speed prediction on segmented road graphs \cite{liu2020graphsage}.

Recent studies have demonstrated the effectiveness of spatio-temporal graph convolutional architectures such as ST-GCN, which model both topological and temporal dependencies in traffic flow data \cite{rahmani2023graph}. Building on this foundation, hybrid models like DHGNN incorporate both heterogeneous data sources and multi-relational sub-networks for improved traffic incident prediction \cite{tran2021data}, while DSTGCN expands the modeling capability by capturing diffusion-based spatial-temporal dependencies in traffic dynamics \cite{rahmani2023graph}. These models often combine GCNs or GATs with recurrent units like GRU or LSTM to forecast demand in ride-hailing services and long-term traffic conditions. A more recent advancement, STZITD-GNN, integrates GAT-based encoders with a zero-inflated Tweedie distribution decoder to effectively manage imbalanced and uncertain crash risk prediction tasks at the road level \cite{gao2024uncertainty}.

However, despite these advancements, current GNN applications in transportation remain largely limited to flow prediction tasks and seldom extend to event-level safety analysis. Gao et al. \cite{gao2024uncertainty} addressed this gap by introducing a probabilistic GNN model for road-level crash risk prediction, incorporating Bayesian uncertainty to enhance model reliability for safety-critical applications. Still, the focus on crash severity classification remains underexplored in most GNN-based frameworks \cite{li2024graph}.

Moreover, GNNs have only recently begun integrating unstructured data such as textual narratives or public sentiment. Zhang et al. \cite{zhang2022multimodal} proposed a multimodal coupled graph attention network that jointly analyzes traffic event texts and sentiment data to enhance incident detection, highlighting the potential of combining semantic and spatial information. Studies by Sun et al. \cite{sun2023attention} and Varhatis et al. \cite{vrahatis2024graph} further support the versatility of attention-based GNNs in fusing diverse data modalities, although such integration is still nascent in transportation safety research. Additionally, while grid-based spatial partitioning methods are commonly used in traffic forecasting \cite{tang2024spatio}, the implementation of hierarchical or multi-resolution graph structures, such as those using coarse regional aggregation like H3 grids, is largely absent from current GNN literature. Li et al. \cite{li2024graph} emphasize this as a promising direction for improving spatial generalization across varying geographic scales. These gaps collectively underscore the need for a more comprehensive GNN framework that can integrate fine- and coarse-level spatial features, encode temporal patterns, and incorporate unstructured crash-related text, motivating the design of our proposed ST-GraphNet model.

\section{Dataset and Preprocessing}
\subsection{Data Source and Description}
This study utilizes crash data for automation-involved vehicles across the state of Texas for the year 2024, obtained from the Texas Crash Records Information System (CRIS). The dataset includes key variables such as spatial coordinates, temporal attributes, SAE automation levels, crash severity outcomes, and narrative descriptions. These narratives provide critical context regarding the circumstances of each crash, enhancing the dataset's utility for both structured and semantic modeling.

Crash severity is categorized using the standard KABC scale, which includes \textit{Killed, Incapacitating Injury, Non-Incapacitating Injury,} and \textit{Possible Injury}, while \textit{Not Injured} represents the no-injury category. Initially, the dataset exhibited a class imbalance, with 2,789 no-injury cases and 1,176 injury cases. To ensure balanced representation for graph modeling, the no-injury class was undersampled to 1,176 records, resulting in a final dataset of 2,352 observations with an equal distribution across both classes.

The dataset also contains detailed crash narratives, which offer additional insights into crash dynamics and contributing factors. For instance, one injury-class narrative (SAE Level 1) states: \textit{``Unit 2 was idled at the stop sign at the intersection of Ebony St \& Nolana Loop, facing southbound. Unit 1 was idled behind Unit 2. Unit 1 failed to control speed thus with its front distributed struck Unit 2 back distributed.''} This description indicates that the crash occurred at an intersection and involved a failure to control speed by the trailing vehicle, resulting in a rear-end collision. The presence of a stationary lead vehicle at a stop sign highlights factors commonly associated with injury outcomes.

In contrast, a no-injury case from SAE Level 2 is described as follows: \textit{``Unit 2 was stationary in the northbound lane at the red light of the 600 block of S. Texas Blvd. Unit 1 was traveling north at the 600 block of S. Texas Blvd. Unit 1 failed to control speed and struck Unit 2 on the back end. Unit 1 sustained damage to the front end.''} While this incident also involves a rear-end collision due to a failure to control speed, the impact likely occurred at a lower velocity or more controlled conditions, contributing to the absence of reported injuries.

All SAE autonomy levels are retained without aggregation to preserve the granularity of automation-level specific crash outcomes. This comprehensive dataset, integrating structured spatio-temporal attributes with unstructured textual narratives, forms the foundation for developing a graph-based neural network framework capable of modeling crash severity through both geometric and semantic reasoning.

\subsection{Crash Narrative Embedding with Language Models}
To encode the semantic information from crash narrative descriptions, Sentence Bidirectional Encoder Representations from Transformers (Sentence-BERT) (\texttt{all-MiniLM-L6-v2}), a pretrained transformer-based language model, has been employed to convert the textual narratives into dense vector embeddings. Each narrative was encoded into a 384-dimensional vector, resulting in a final embedding matrix of shape $(3965, 384)$, which was incorporated as part of each node’s feature vector in the graph.

\section{Graph Construction}
\subsection{Fine‐Grained Graph Construction}
\subsubsection{Node Definition and Features}

In the fine‐grained graph, each node corresponds to a single crash event. After balancing and preprocessing the raw crash dataset, every retained record is associated with a unique node index. The feature vector for each node combines both numeric attributes and semantically rich text embeddings. Numerically, each crash record contributes its SAE autonomy level (an integer category from 0 to 5) as well as two pairs of cyclical temporal features: \(\sin\) and \(\cos\) encodings for both the hour of day and the weekday. To capture temporal context, the \texttt{CrashDate} and \texttt{CrashTime} attributes are combined into a single \texttt{timestamp} field and extracted both the hour of the day and the day of the week. To preserve the cyclical nature of these time-related features, they applied sine and cosine transformations to the hour and weekday variables. These cyclical encodings ensure that temporal continuity (e.g.\ midnight wrapping to hour 0) is preserved in a continuous vector space. Concretely, if a crash occurred at hour \(h\) (0 through 23), the feature pair is computed.

\begin{equation}
\begin{aligned}
    \text{hour\_sin} &= \sin\left(2\pi \cdot \frac{\text{h}}{24}\right), \\
    \text{hour\_cos} &= \cos\left(2\pi \cdot \frac{\text{h}}{24}\right)
\end{aligned}
\end{equation}

 Similarly, for the weekday \(d\) (0 = Monday through 6 = Sunday), the pair is appended. By stacking these four temporal dimensions together with the single numeric SAE level, a total of five numeric features represent each crash’s discrete and cyclic attributes.

\begin{equation}
\begin{aligned}
    \text{weekday\_sin} &= \sin\left(2\pi \cdot \frac{\text{d}}{7}\right), \\
    \text{weekday\_cos} &= \cos\left(2\pi \cdot \frac{\text{d}}{7}\right)
\end{aligned}
\end{equation}

To infuse contextual, unstructured information from the crash narrative into each node, a pretrained Sentence‐BERT model (“all‐MiniLM‐L6‐v2”) is applied to convert every narrative string into a 384-dimensional embedding. These embeddings capture latent semantic patterns such as mentions of road conditions or participant activities enabling the graph neural network to leverage both structured (numeric) and unstructured (textual) signals. After computing the 384-dimensional embedding for each narrative, the feature extraction pipeline horizontally concatenates the five numeric values with the 384-dimensional vector, producing a single 389-dimensional feature vector for each node. Finally, each node is labeled by the binary severity outcome (0 = no injury, 1 = injury) derived from the original crash severity categories. These labels serve as targets for downstream node-level classification tasks.

\subsubsection{Spatio‐Temporal Edge Definition}

Edges in the fine‐grained graph encode both spatial proximity and temporal closeness between crash events. To capture spatial relationships, the Haversine formula is employed to compute great‐circle distances on the Earth’s surface. Given two crash locations \((\mathit{lat}_i,\mathit{lon}_i)\) and \((\mathit{lat}_j,\mathit{lon}_j)\), the Haversine function returns the distance in kilometers. Two nodes \(i\) and \(j\) are considered spatially adjacent if and only if the Haversine distance between their recorded GPS coordinates does not exceed \(30\,\text{km}\). Temporally, an edge is permitted only if the absolute time difference between the two crashes is at most \(24\) hours. This is implemented by casting the \texttt{timestamp} column to NumPy’s \texttt{datetime64[ns]} dtype and computing
\[
\bigl\lvert \text{time}_j - \text{time}_i \bigr\rvert \,/\, \texttt{np.timedelta64(1,\, 'h')}.
\]
If this computed duration is \(\le 24\) hours, the pair passes the temporal criterion.

In practice, the code loops over all ordered crash pairs \((i, j)\) with \(j > i\). For each pair, it first checks the temporal difference; only if 
\[
\lvert \text{time}_j - \text{time}_i \rvert \le 24 
\]
does it then compute the Haversine distance. If the distance is at most \(30\,\text{km}\), two directed edges \((i \rightarrow j)\) and \((j \rightarrow i)\) are appended to the edge list to model an undirected connection in the graph. After processing every crash pair, the accumulated edge list is converted into a \([2\times E]\) PyTorch tensor \(\texttt{edge\_index}\), where \(E\) is the final count of directed edges. The resulting fine-grained \(\texttt{Data}\) object contains \(\mathbf{x}\) (the \([N\times 389]\) feature matrix), \(\texttt{edge\_index}\) (the \([2\times E]\) adjacency tensor), and \(\mathbf{y}\) (the length-\(N\) label vector). This construction ensures that information propagates across both nearby crashes in space and those occurring within a one-day window, enabling the graph neural network to learn joint spatio-temporal patterns underlying injury outcomes.

\begin{figure}[ht]
  \centering
  \includegraphics[width=\linewidth]{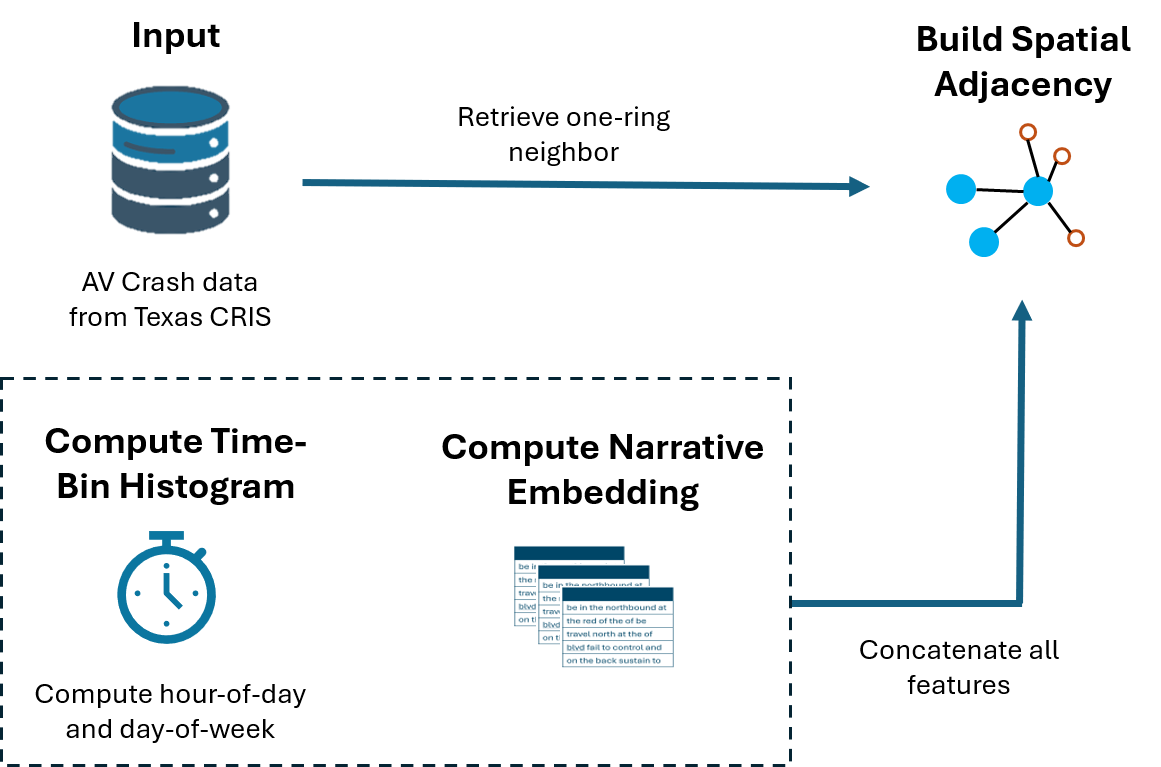}
  \caption{Fine Grain Graph Architecture}
  \label{fig:dstgcn_arch}
\end{figure}

\subsection{Coarse-Grained Graph Construction (H3 Grid)}

\subsubsection{Hexagonal Aggregation and Connectivity}

The coarse‐grained graph is constructed by first mapping each individual crash event into a uniform hexagonal grid using Uber’s Hexagonal Hierarchical Spatial Indexing (H3). At resolution 7, H3 partitions the study area into hexagons of approximately 5.16 km\textsuperscript{2} each, ensuring that every cell has roughly equal area and a consistent six‐neighbor adjacency. After verifying that each crash record contains non‐missing latitude and longitude values, the H3 index for each crash is computed with \texttt{h3.latlng\_to\_cell(latitude, longitude, 7)}. Crashes that share the same H3 index are grouped into a single hexagonal region. Once all occupied hexagons have been identified and transformed into nodes, edges are defined by consulting H3’s built‐in adjacency information. For each hexagon, its one‐ring neighbors—those sharing a contiguous hexagonal boundary—are retrieved using \texttt{h3.grid\_disk(h3\_index, 1)}. Whenever a neighboring hexagon also contains at least one crash (i.e., appears in the set of nodes), a bidirectional edge is created between the two corresponding node indices. These directed edge pairs are collected and converted into a PyTorch tensor of shape \([2, E]\), where \(E\) is the total number of directed edges. In this way, the hexagon grid’s uniform topology is preserved, and spatial information can propagate across adjacent regions during graph neural network training.

\subsubsection{Node Feature Aggregation}

Each H3 cell node aggregates multiple modalities of information from the crashes it contains. First, because Vehicle Industry SAE autonomy levels are discrete categories (ranging from 0 to 5), a histogram of autonomy‐level counts is computed: each bin records how many crashes of a particular SAE level occurred within that hexagon. Second, \texttt{Crash\_Severity} labels are converted into a binary variable (0 = no injury, 1 = injury), and the total counts of non‐injury versus injury crashes in each cell are tallied, producing a two‐dimensional severity‐count vector. Third, temporal patterns are captured by constructing time‐bin histograms: the number of crashes occurring in each hour of day (24 bins) and in each day of week (7 bins) are counted. Fourth, each crash narrative is embedded into a 384‐dimensional vector using a pretrained Sentence‐BERT model; these embedding vectors are averaged across all crashes in the same cell, yielding a single semantic summary that reflects the predominant textual content of narratives in that region. By concatenating the SAE histogram, binary severity counts, hour‐of‐day histogram, weekday histogram, and averaged narrative embedding, a comprehensive feature vector is obtained for every hexagon node.

These feature vectors are then stacked into a tensor of shape \([N, F]\), where \(N\) is the number of hexagon nodes and \(F\) equals the combined dimension of SAE histogram (number of distinct autonomy levels), severity counts (2), hour histogram (24), weekday histogram (7), and narrative embedding (384). Finally, each node’s label is assigned according to the majority of binary severity values within its hexagon, with “injury” selected if injury‐level crashes outnumber non‐injury crashes. The resulting node features, edge index, and node labels are encapsulated in a \texttt{torch\_geometric.data.Data} object named \texttt{coarse\_graph}. By grouping raw crash points into hexagonal regions, summarizing each region’s autonomy‐level distribution, severity frequency, temporal patterns, and narrative semantics, and linking adjacent hexagons via H3’s uniform grid, this coarse‐grained graph offers an efficient, region‐level representation that preserves essential spatial, temporal, and multi‐modal information for downstream graph neural network training. 

\begin{figure}[ht]
  \centering
  \includegraphics[width=\linewidth]{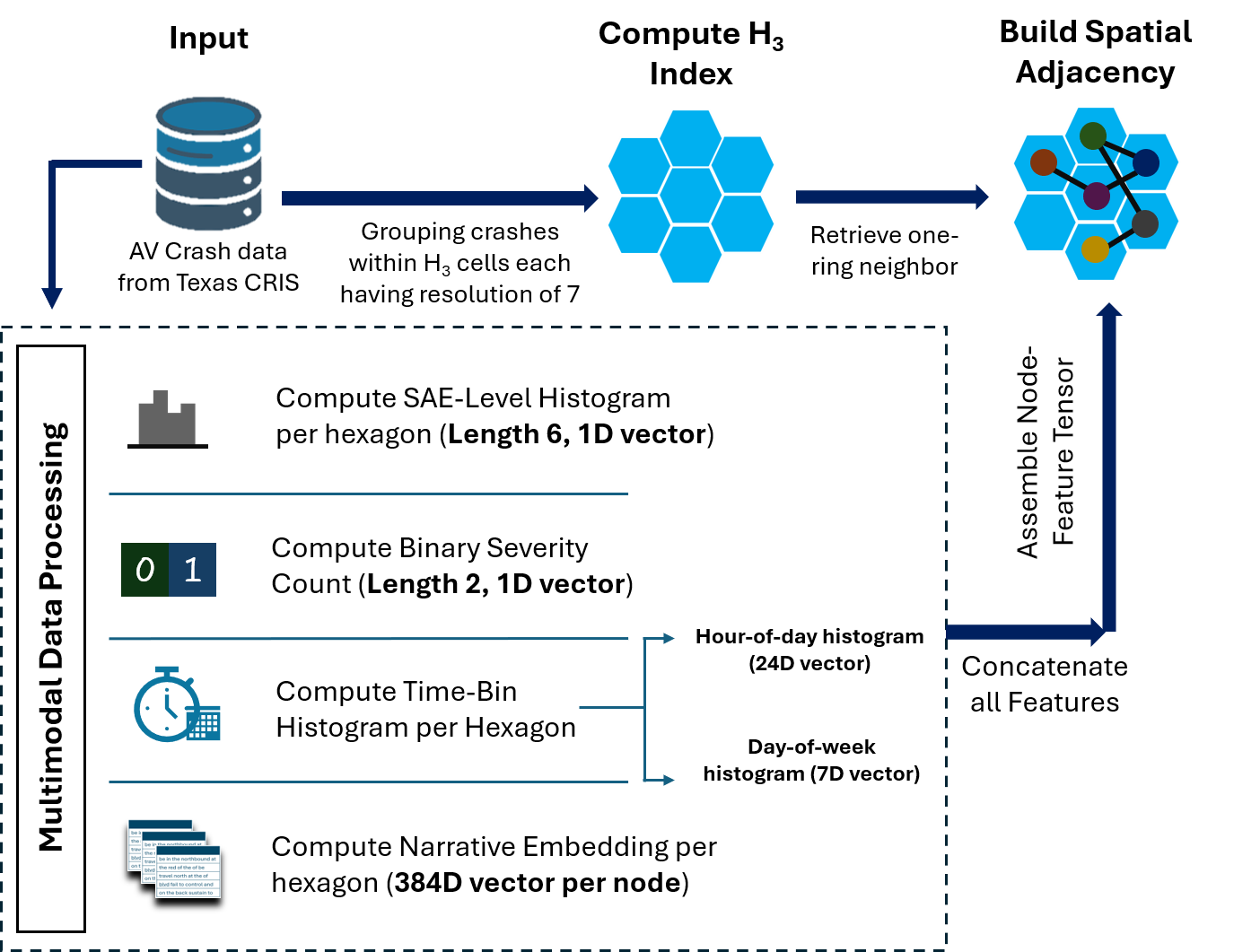}
  \caption{Coarse Grain Graph Architecture}
  \label{fig:dstgcn_arch}
\end{figure}

\begin{figure*}[ht]
  \centering
  \includegraphics[width=0.8\textwidth]{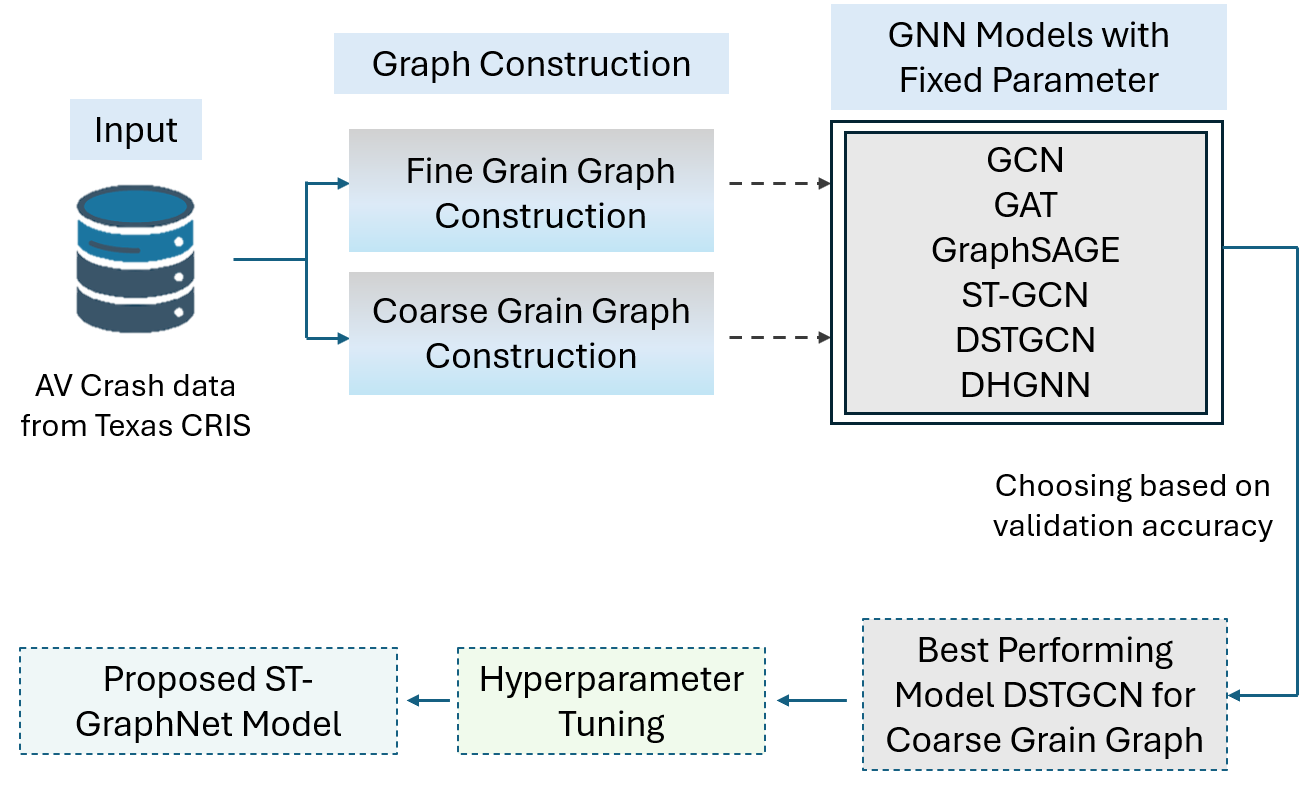}
  \caption{Proposed Study Design}
  \label{fig:roc_curves}
\end{figure*}

\section{Experimental Setup}

\subsection{Task Definition and Evaluation Metrics}

The primary task addressed in this study is the node‐level classification of crash severity within a spatio‐temporal graph constructed from AV crash data. Each node represents either an individual crash event (in the fine‐grained graph) or an aggregated hexagonal region (in the coarse‐grained graph), and is characterized by multi‐modal attributes including numerical features (e.g., SAE autonomy level, encoded hour‐of‐day and day‐of‐week signals), 384‐dimensional narrative embeddings derived from crash descriptions, and when applicable-time‐bin histograms capturing hourly or weekly crash frequencies. Spatial connectivity is encoded via Haversine‐based adjacency for fine‐grained graphs (connecting crashes that occur within 30 km and 24 hours of one another) or via H3 hexagonal adjacency for coarse‐grained graphs (linking adjacent hexagonal cells). In models that explicitly incorporate temporal dependencies, additional layers process the temporal evolution of node states across successive time bins.

The objective is to predict, for each node, a binary severity label indicating whether a crash resulted in injury (1) or no injury (0) using multiple GNN models and selecting the best one among them. During training and evaluation, the dataset is partitioned into training, validation, and test masks (e.g., 70 \%20 \%10 \% splits) to ensure that the learned representations generalize to unseen nodes. Performance is primarily assessed using metrics suited for binary classification: overall accuracy, weighted F1‐score, and confusion matrix statistics (true positives, false positives, true negatives, and false negatives). The weighted F1‐score is particularly critical, as it balances precision and recall across both “injury” and “no‐injury” classes, thus providing a robust measure of model effectiveness. 

To identify the final model, the validation F₁‐score served as the primary selection criterion. Once the leading architecture was chosen, its hyperparameters were exhaustively tuned to maximize validation performance. Subsequently, several complementary evaluation metrics were computed for the top‐performing configurations. In particular, receiver operating characteristic (ROC) curves and their corresponding area under the curve (AUC) values were calculated to assess the model’s ability to differentiate between injury and non‐injury classes across varying probability thresholds. Precision–recall curves, together with average precision (AP) scores, were also generated, offering a targeted measure of performance in the context where correctly identifying injury cases is paramount. Together, these evaluation metrics quantify how well the proposed spatio‐temporal graph neural network (ST‐GraphNet) architectures leverage multi‐modal node features, spatial structure, and temporal dynamics to produce accurate and reliable crash severity predictions.

\subsection{Model Training and Hyperparameters}
The initial hyperparameter configuration was selected to establish a consistent foundation for comparing diverse graph neural network (GNN) architectures on both fine‐grained and coarse‐grained crash‐severity prediction tasks. The evaluated architectures included Graph Convolutional Network (GCN), Graph Attention Network (GAT), Graph Sample and Aggregated Embeddings (GraphSAGE), Spatio‐Temporal Graph Convolutional Network (ST‐GCN), Dynamic Spatio‐Temporal Graph Convolutional Network (DSTGCN), and Dynamic Hierarchical Graph Neural Network (DHGNN). To ensure that differences in performance could be ascribed to model design rather than training settings, every network was trained with the same set of hyperparameters (as shown in Table \ref{tab:hyperparams}): a hidden‐layer dimension of 32, a dropout rate of 0.30, and an Adam optimizer learning rate of 0.05. An L2 weight‐decay of 0.005 was applied to each model to mitigate overfitting, and training was conducted for 30 epochs, sufficient to achieve stable validation metrics without incurring excessive computation time. Finally, the dataset was partitioned so that 70\% of the instances were used for training, 20\% for validation, and the remaining 10\% for testing. By holding these hyperparameters constant across all architectures, any observed variation in validation performance could be attributed directly to the relative strengths and weaknesses of the GNN designs themselves.

\begin{table}[ht]
  \centering
  \small
  \setlength\tabcolsep{4pt} 
  \caption{Hyperparameter Settings}
  \label{tab:hyperparams}
  \begin{tabular}{l p{4cm} c}
    \toprule
    \textbf{Hyperparameter} & \textbf{Description} & \textbf{Value} \\
    \midrule
    \texttt{hid\_dim}       & Hidden-layer size                        & 32     \\
    \texttt{dropout}        & Dropout probability                      & 0.30   \\
    \texttt{learning\_rate} & Adam optimizer learning rate             & 0.05   \\
    \texttt{weight\_decay}  & Adam optimizer weight-decay (L2 regularization)    & 0.005  \\
    \texttt{num\_epochs}    & Number of training epochs                & 30     \\
    \bottomrule
  \end{tabular}
\end{table}

\subsection{Baselines and Final Model Selection}
In examining the performance of various GNN architectures on the task of predicting crash severity, a clear distinction emerged between the fine‐grained and coarse‐grained graph constructions as shown in Table~\ref{tab:val_f1_comparison}. Under the fine‐grained paradigm, where each crash event is represented as an individual node and edges are created only between events occurring within 30 km and 24 hours, none of the tested models achieved robust validation scores. For example, a basic GCN achieved only a 0.6079 validation F1, while a GAT reached 0.5958. The spatio‐temporal variant ST‐GCN fared worse, with a validation F1 of just 0.5148. Even more sophisticated approaches, such as GraphSAGE (0.5845) and DSTGCN (0.6164), could not surpass the 0.65 mark, and DHGNN, which incorporates hierarchical attention, achieved a peak of 0.6441. These results indicate that when each crash stands alone, connected only to a handful of temporally and spatially proximate peers, the signal is too noisy and sparse for reliable crash severity classification.

Aggregation into a coarse-grained representation via H3-indexed hexagonal cells (each ~5.16,km\textsuperscript{2}) enriched region-level context by summarizing SAE autonomy-level histograms, binary injury counts, averaged narrative embeddings, and time-bin histograms, thereby reducing noise. Within this framework, GCN and GAT each achieved a validation F1 of 0.9098; GraphSAGE reached 0.9775; and both DSTGCN and DHGNN attained 0.9850. The uniform six-neighbor H3 adjacency enabled effective exploitation of spatial and semantic correlations. This region-level aggregation transforms diffuse crash records into a structured, information-rich graph in which even simpler GNNs can deliver high accuracy.

Coarse-graining into H3 hexagons boosted validation F1 from ~0.6 to ~0.98 across all models (for DSTGCN: 0.6164→0.9850; for DHGNN: 0.6441→0.9850; for GraphSAGE: 0.5845→0.9775), prompting exclusive focus on the coarse-grained graph. Although both DSTGCN and DHGNN achieved 0.9850, DSTGCN’s simpler linear + convolutional architecture was chosen as the primary candidate. The next phase involves tuning learning rate, dropout, hidden-dimension size, and time-bin resolution to finalize ST-GraphNet, followed by benchmarking against established methods.

This comparative analysis demonstrates two key findings. First, the coarse‐grained, H3‐based aggregation produces better validation results across all GNN architectures, effectively transforming the problem into one that can be modeled with high confidence. Second, among the evaluated models on the coarse graph, DSTGCN and DHGNN exhibit the strongest validation performance. Given DSTGCN’s relative simplicity and ease of interpretation, it is chosen for subsequent fine‐tuning. The final tuned DSTGCN (i.e., ST-GraphNet) will then serve as a robust, interpretable framework for predicting crash severity in automated vehicle datasets.```

\begin{table}[ht]
  \centering
  \caption{Best Validation F1 Scores: Fine‐Grained vs. Coarse‐Grained Graphs}
  \label{tab:val_f1_comparison}
  \resizebox{\linewidth}{!}{%
    \begin{tabular}{lcc}
      \toprule
      \textbf{Model}    & \textbf{Fine‐Grained Best Val F1} & \textbf{Coarse‐Grained Best Val F1} \\
      \midrule
      GCN               & 0.6079                             & 0.9098                              \\
      GAT               & 0.5958                             & 0.9098                              \\
      GraphSAGE         & 0.5845                             & 0.9775                              \\
      ST‐GCN            & 0.5148                             & 0.8861                              \\
      DSTGCN            & 0.6164                             & 0.9850                              \\
      DHGNN             & 0.6441                             & 0.9850                              \\
      \bottomrule
    \end{tabular}%
  }
\end{table}

\section{Proposed Method: ST-GraphNet Framework}
\subsection{Training and Fine-tuning ST-GraphNet Model}
To identify the optimal configuration for ST-GraphNet, the DSTGCN was subjected to an extensive hyperparameter search over key architectural and optimization parameters. In particular, the grid encompassed variations in hidden-layer dimension (32 vs.\ 64), dropout probability (0.3 vs.\ 0.4), learning rate (0.10, 0.07, 0.04, 0.001), and L2 weight-decay ($5\times10^{-3}$, $5\times10^{-4}$, $5\times10^{-5}$). For each candidate setting, the coarse-grained H3 graph was split into 70 \% of nodes for training, 20 \% for validation, and 10 \% for final testing. To ensure comparability across all grid configurations, the number of epochs was fixed at 30.

\begin{figure}[ht]
  \centering
  \includegraphics[width=0.5\textwidth]{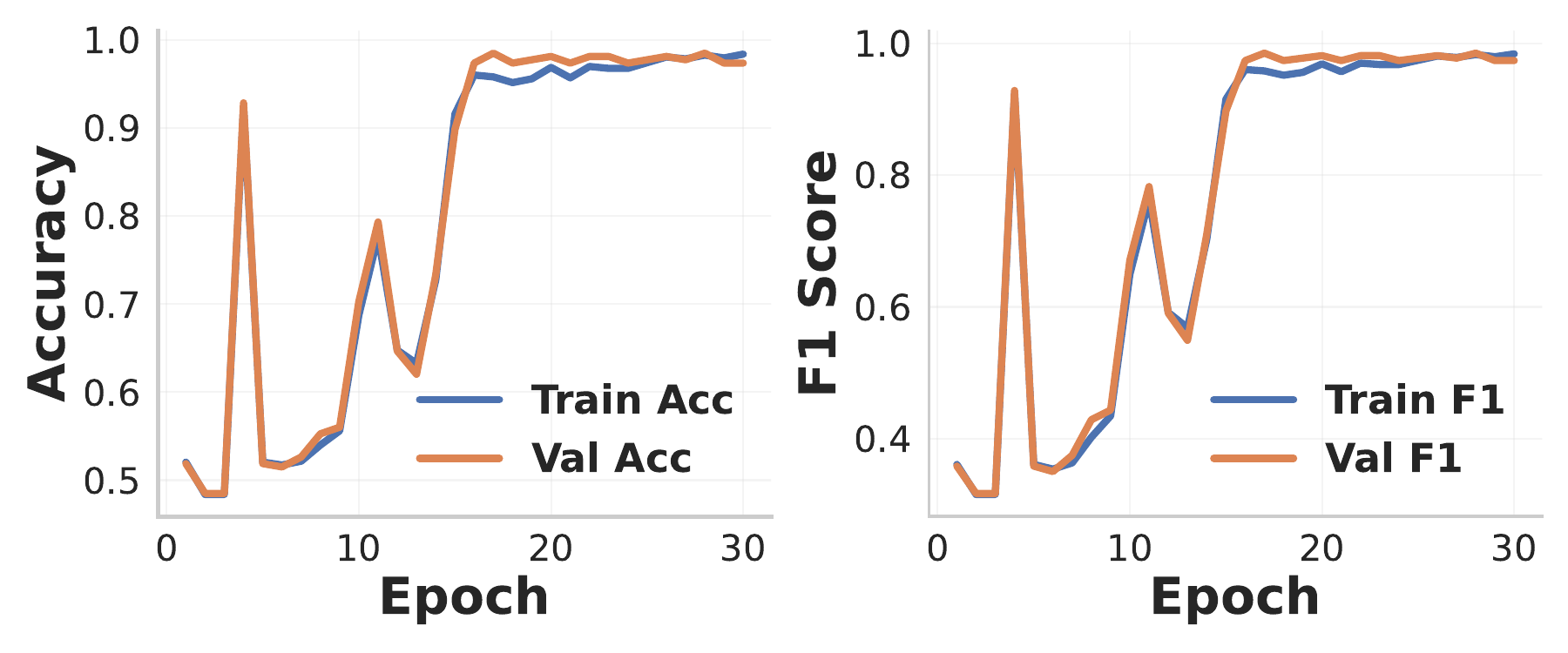}
  \caption{Left: Training and validation accuracy over 30 epochs. Right: Training and validation F1 score over 30 epochs. Notice how both metrics improve rapidly and the small gap between train/val indicates overfitting is controlled.}
  \label{fig:train_val_metrics}
\end{figure}

After evaluating every combination, the best validation F1 score (0.9850) was obtained using a DSTGCN with a hidden-layer size of 64, dropout 0.30, learning rate 0.07, and weight-decay 0.0005. Under this configuration, the training set comprised 928 hexagon nodes, and the validation set comprised 266 nodes. During training, ST-GraphNet demonstrated rapid convergence in both accuracy and weighted F1. Specifically, validation accuracy progressed from near-random levels in early epochs to approximately 0.9797 at epoch 30, while the corresponding training accuracy reached 0.9694 as shown in Figure~\ref{fig:train_val_metrics}. Similarly, the validation F1 score steadily improved, ultimately peaking at 0.9850, reflecting a balanced precision–recall tradeoff for the binary ``Injury'' versus ``Not Injured'' classification task. Crucially, the small gap (\(\approx 0.01\text{--}0.02\)) between training and validation metrics throughout the 30-epoch run indicated that overfitting was effectively controlled by the chosen dropout and weight-decay settings.

By fixing all other hyperparameters and selecting this top performing DSTGCN configuration, ST-GraphNet establishes a robust foundation for downstream testing. In subsequent sections, the trained weights corresponding to the validation optimum (F1 = 0.9850) are evaluated on the held-out test set to demonstrate generalization performance. These results confirm that the selected hyperparameters achieve an optimal balance between model capacity and regularization, allowing ST-GraphNet to utilize coarse-grained H3 aggregation, multi-modal node features, and spatio-temporal dynamics to generate highly accurate crash severity predictions.

\subsection{Overview of ST-GraphNet Architecture}
The proposed \textbf{ST-GraphNet} framework integrates a coarse-grained spatio-temporal graph with the DSTGCN \cite{chen2019dstgcnn} to predict AV crash severity from multimodal crash data. The architecture captures both spatial correlations across node-level aggregates and temporal evolution patterns. Our proposed model, ST-GraphNet, is built upon a DSTGCN backbone, customized to operate on an H3‐aggregated hexagon graph. Each node in the graph represents a 5.16 km\textsuperscript{2} hexagonal region containing multiple crash events, and is endowed with a multi‐modal feature vector comprising: (1) an SAE‐level histogram (six bins for autonomy levels 0–5), (2) binary counts of injury vs.\ no‐injury crashes, (3) a 24‐bin hour‐of‐day histogram, (4) a 7‐bin weekday histogram, and (5) a 384‐dimensional averaged narrative embedding from Sentence‐BERT. These combined features form an \(F\)-dimensional input vector for each node, where \( [F = 6 + 2 + 24 + 7 + 384 = 423]\).

The H3 adjacency defines a uniform six-neighbor hexagon grid: two nodes are connected by an undirected edge if and only if their corresponding hexagon indices share a common boundary. This adjacency is encoded as a sparse edge index tensor \(\texttt{edge\_index}\in \mathbb{N}^{2 \times E}\) for use with PyTorch Geometric. In this coarse‐grained representation, the spatial graph remains static throughout training.

ST-GraphNet’s forward pass consists of \(L\) stacked DSTGCN layers. Each DSTGCN layer integrates two components: a graph convolution operation to diffuse information across neighboring hexagons, and a temporal convolution (1D) that captures sequential dependencies among node feature channels. 

Concretely, let \(\mathbf{H}^{(0)} \in \mathbb{R}^{N \times F}\) denote the input feature matrix, where \(N\) is the number of hexagon nodes. In layer \(l\), the layer first applies a spatial graph convolution:
\[
\widetilde{\mathbf{H}}^{(l)} \;=\; \sigma\Bigl(\widehat{\mathbf{A}}\,\mathbf{H}^{(l-1)}\,\mathbf{W}^{(l)}_{\text{spatial}} \Bigr),
\]
where \(\widehat{\mathbf{A}} = \mathbf{D}^{-1/2} (\mathbf{A} + \mathbf{I}) \mathbf{D}^{-1/2}\) is the renormalized adjacency (with self-loops), \(\mathbf{W}^{(l)}_{\text{spatial}}\in \mathbb{R}^{F_{l-1} \times F_{l}}\) is a learnable weight matrix, \(F_{l-1}\) is the input dimension to layer \(l\), \(F_{l}\) is the output dimension of layer \(l\), and \(\sigma\) is a nonlinear activation (typically ReLU). Immediately following the spatial convolution, ST-GraphNet applies a temporal convolution (1D) across the feature dimensions of each node:
\[
\mathbf{H}^{(l)} \;=\; \sigma\Bigl(\text{Conv1D}\bigl(\widetilde{\mathbf{H}}^{(l)}; \;\mathbf{W}^{(l)}_{\text{temporal}}\bigr)\Bigr),
\]
where \(\mathbf{W}^{(l)}_{\text{temporal}}\) is a 1D convolution kernel of size \(k_t\) (e.g., \(k_t=3\)) that slides across the feature channels for each node independently, thereby modeling how patterns in SAE levels, temporal histograms, and narrative semantics evolve jointly. Between each DSTGCN layer, a dropout layer (with rate 0.30) is inserted to mitigate overfitting by randomly zeroing a fraction of the features. The hidden‐layer dimension is kept constant at 64 for each intermediate layer.

After \(L\) DSTGCN blocks (where \(L=2\) or \(3\) depending on the chosen depth), the final node‐level representations \(\mathbf{H}^{(L)} \in \mathbb{R}^{N \times F_{L}}\) are pooled through a linear classification head:
\[
\mathbf{Z} \;=\; \text{softmax}\Bigl(\mathbf{H}^{(L)}\,\mathbf{W}_{\text{out}} + \mathbf{b}_{\text{out}}\Bigr),
\]
where \(\mathbf{W}_{\text{out}}\in \mathbb{R}^{F_{L} \times 2}\) and \(\mathbf{b}_{\text{out}}\in \mathbb{R}^{2}\) project each node representation to a 2-dimensional logit vector \((z_{i,0}, z_{i,1})\), corresponding to the probabilities of “no injury” (class 0) and “injury” (class 1). A cross-entropy loss is computed over the training nodes to update all weight matrices via Adam optimization with a learning rate of 0.07 and weight‐decay of \(5\times10^{-4}\).

ST-GraphNet first encodes each node by fusing SAE‐level and temporal histograms with narrative embeddings, then applies a renormalized graph convolution over the six‐neighbor hexagon mesh to ensure stable, topology‐aware message passing. Next, temporal convolutions run along the stacked feature channels—rather than across separate time‐step graphs—to capture cross-modal correlations directly. Finally, a lightweight linear head projects the learned embeddings into two logits, with softmax producing the crash-severity probabilities.

By jointly learning spatial dependencies through graph convolution and temporal/semantic interdependencies through 1D convolutions over feature channels, ST-GraphNet effectively synthesizes heterogeneous input modalities into a unified latent space. This architecture enables the model to detect complex spatio-temporal patterns, such as rush-hour clustering of injury crashes in adjacent hexagons or semantic co-occurrence of “intersection” and “high speed” phrases, that would be difficult to capture using only grid-based or purely temporal methods. Consequently, ST-GraphNet achieves high performance (F1 = 0.9850 on validation; AUC = 0.998 on test) and demonstrates strong potential for guiding AV safety interventions in real urban environments.  

\begin{figure*}[ht]
  \centering
  \includegraphics[width=\textwidth]{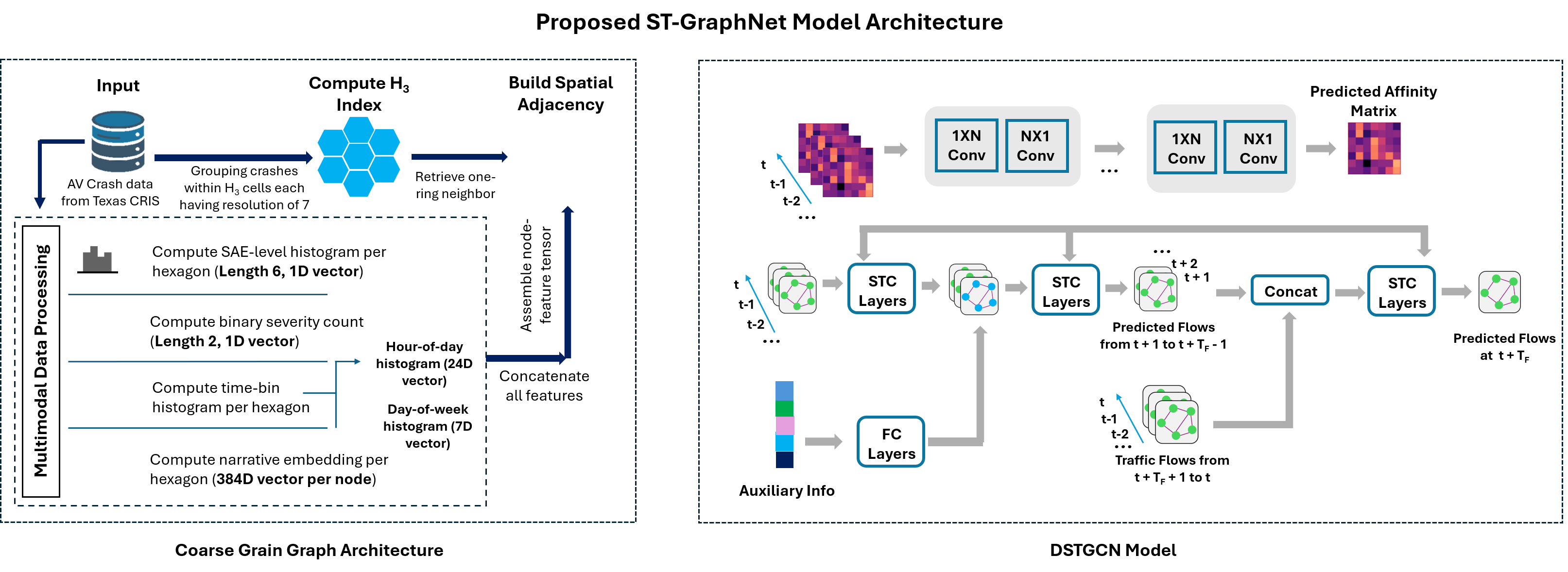}
  \caption{Overview of the proposed ST-GraphNet framework. On the left, raw and contextual inputs are combined in a multimodal data processing block to produce two parallel streams. On the right, the temporal convolution stream (top) applies successive 1 × N and N × 1 convolutions to extract evolving spatial patterns from traffic matrices, while the graph prediction and propagation stream (bottom) dynamically learns time-varying adjacency (affinity) matrices and feeds them into stacked Spatio-Temporal Graph Convolution (STC) layers.}
  \label{fig:dstgcn_arch}
\end{figure*}

\subsection{ST-GraphNet Model Results}
The ST-GraphNet model demonstrates exceptionally strong performance across all evaluation metrics, indicating both high predictive accuracy and robust generalization to unseen data. When the best validation‐state weights were loaded and the model was evaluated on the held‐out test set, it achieved a test accuracy of 0.9774 and a weighted F1 score of 0.9774. The corresponding confusion matrix on the test, revealed only 3 misclassifications out of 133 test nodes: 62 “Not Injured” crashes were correctly identified, 2 “Not Injured” cases were mislabeled as “Injury,” 1 “Injury” case was mislabeled as “Not Injured,” and 68 “Injury” crashes were correctly classified. These numbers translate to a true negative rate of 0.969, a true positive rate of 0.986, and an overall misclassification rate below 2.3 \%. Such performance indicates that the model is very effective at distinguishing between “Injury” and “Not Injured” outcomes at the coarse‐grained (H3 cell) level.

Beyond point‐estimate metrics, the DSTGCN’s classifier confidence was also examined via per‐class ROC curves. The area under the ROC curve (AUC) for both classes (“Not Injured” and “Injury”) was 0.998, as shown in Figure~\ref{fig:roc_curves}, effectively reaching the maximum possible value of 1.000. Visually, each ROC curve hugs the top‐left corner of the plot, indicating near‐perfect separation between positive and negative instances. Equally telling, the Precision–Recall curve yielded an average precision (AP) score of approximately 0.997 for the positive (“Injury”) class. A high AP score demonstrates that the model maintains both high precision (few false positives) and high recall (few false negatives) even as the decision threshold moves.


\begin{figure}[ht]
  \centering
  \includegraphics[width=0.4\textwidth]{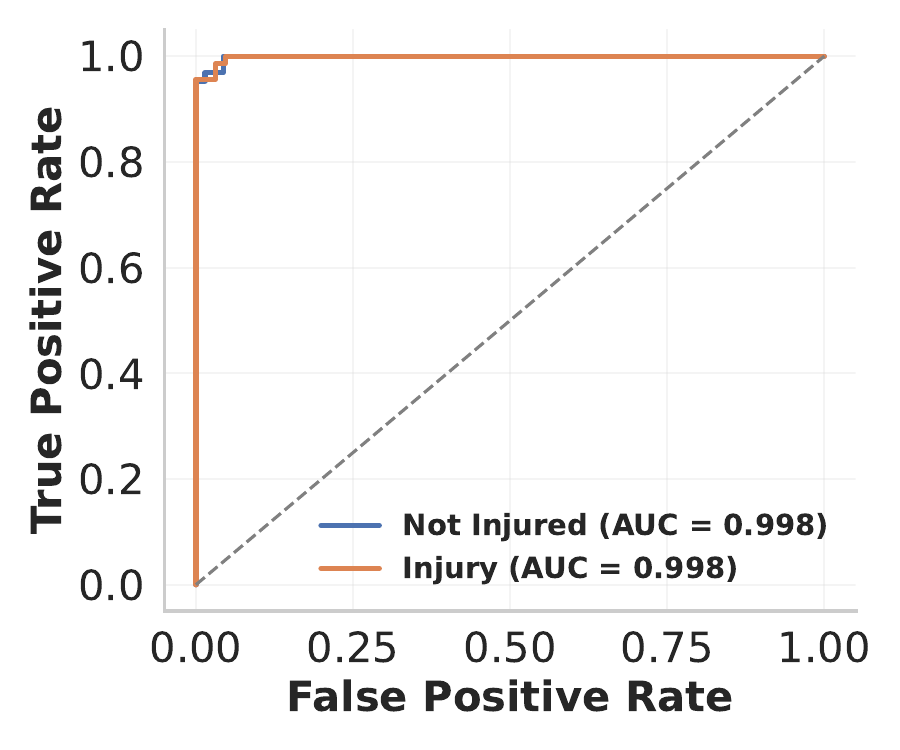}
  \caption{Receiver Operating Characteristic (ROC) curves for each class on the test set. Both “Not Injured” and “Injury” crash classes achieve an AUC of 0.998.}
  \label{fig:roc_curves}
\end{figure}

\section{Discussions}
The ST-GraphNet experiments uncover key insights into how spatio-temporal aggregation and graph modeling can clarify the underlying patterns of AV crash severity. First, the dramatic performance improvement observed when moving from the fine-grained to the coarse-grained H3 hexagon representation indicates that aggregating individual crash events into spatially contiguous regions captures a more stable risk pattern. In the fine-grained graph, each crash node was connected only to events occurring within a 30\,km/24\,h window; as a result, the underlying adjacency was sparse and highly sensitive to individual event noise. In contrast, grouping crashes into H3 cells at resolution 7 (\(\approx 5.16\) km\textsuperscript{2} per hexagon) effectively ``smooths'' local fluctuations: each hexagon node aggregates counts of injuries, counts of no-injury outcomes, SAE-level histograms, temporal histograms (hour-of-day, day-of-week), and averaged narrative embeddings. This richer regional feature set, combined with a consistent six-neighbor topology, allows ST-GraphNet to learn stronger representations of spatial hotspots and temporal trends.

Second, the strong performance of DSTGCN on the coarse-grained graph, with a validation F1 of 0.9850, highlights the importance of modeling temporal dynamics alongside spatial adjacency. By encoding the hourly and weekly crash histograms of each hexagon node into the feature vector, ST-GraphNet employs cyclic temporal signals (e.g., rush-hour peaks, weekday/weekend differences) to distinguish regions that are consistently high-risk from those that experience only transient fluctuations spikes. For example, regions with concentrated late-night or early-morning crash frequencies—often associated with reduced visibility or impaired driving—exhibit distinct sine/cosine signatures that the model can learn. At the same time, the H3 adjacency allows information to diffuse between neighboring cells, capturing how crash‐prone areas tend to cluster spatially (e.g., a busy urban intersection surrounded by several high-traffic segments).

Third, the inclusion of narrative embeddings enriches the learned representations with semantic context that numeric features alone cannot provide. When crash narratives mention contributing factors, such as “failure to yield at red light,” “rear-end collision at 45 mph,” or “loss of control due to wet pavement”, the Sentence-BERT embeddings encode these patterns in a continuous 384-dimensional space. During training, ST-GraphNet learns to associate certain textual clusters (e.g., descriptions mentioning “intersection,” “braking too late,” “rear-end”) with higher injury probabilities. In practice, hexagon nodes whose aggregated narratives describe high-speed impacts or complex intersection maneuvers show elevated node representations in the learned latent space, helping the model separate them from regions where crashes more often occur under lower‐risk conditions (e.g., parking maneuvers or low-speed maneuvering).

These insights demonstrate that a multi-modal, spatio-temporal graph captures latent risk factors that single-modality or purely grid-based approaches would likely miss. The coarse-grained H3 aggregation reduces noise and enforces topological consistency, while dynamic message passing (as in DSTGCN) capitalizes on both spatial adjacency and temporal sequences. Finally, semantic embeddings from crash narratives allow ST-GraphNet to incorporate unstructured domain knowledge, resulting in highly accurate injury-severity predictions.

\section{Conclusions}
In this study, ST-GraphNet was introduced as a novel spatio-temporal GNN framework that leverages coarse-grained H3 hexagon aggregation, multi-modal node features (SAE histograms, temporal histograms, and narrative embeddings), and dynamic message passing to predict AV crash severity with exceptional accuracy (test F1 = 0.9774, AUC = 0.998). By demonstrating that hexagon-based aggregation reduces noise and improves signal quality compared to a fine-grained approach, it shows how region-level representations allow the model to identify spatial hotspots and temporal trends more effectively. 

Despite its high predictive accuracy, ST-GraphNet exhibits certain limitations that warrant consideration. First, the model’s reliance on Texas-specific AV crash data from a single year means that learned patterns may not generalize directly to other regions or time periods, as roadway geometries, enforcement practices, and reporting conventions can vary substantially. Second, the choice of H3 resolution (5.16 km\textsuperscript{2} per hexagon) provides a useful balance between spatial granularity and noise reduction, but alternative resolutions could either obscure local hotspots (if coarsened) or reintroduce data sparsity (if refined). Third, the static graph construction, which aggregates events over an entire year, cannot capture real-time shifts in risk (e.g., temporary construction zones or sudden weather changes) unless the model is retrained frequently. Finally, while ST-GraphNet’s performance is impressive, its deep-learning architecture remains a black box; extracting clear causal explanations (such as “rear-end collisions at high speed on wet pavement drive injury risk”) requires additional interpretability modules, and without them, decision-makers may find it difficult to trace specific interventions back to the model’s internal reasoning.

Looking forward, the proposed ST-GraphNet can be extended along several directions: (1) incorporating multi-resolution hexagon layers so that fine- and coarse-scale features are fused dynamically; (2) integrating real-time traffic and weather feeds to detect emergent risk surges; (3) exploring online, streaming GNN architectures to update node features and adjacency incrementally without full retraining; (4) developing post-hoc explainability methods—such as attention visualizations or gradient-based saliency maps—to make model decisions more transparent; and (5) validating the framework on AV crash data from other states and years to assess generalizability. These enhancements will transform ST-GraphNet into a more versatile, interpretable decision-support tool for proactive AV safety management.


\bibliographystyle{ACM-Reference-Format}
\bibliography{sample-base}










\end{document}